\documentclass[utf8]{frontiersFPHY}

\setcitestyle{square}
\usepackage{url,hyperref,lineno,microtype,subcaption}
\usepackage[onehalfspacing]{setspace}
\usepackage{color}

\newcommand{\FB}[1]{{#1}}
\newcommand{\PA}[1]{{#1}}

\newcommand{\modPB}[1]{{#1}}

\def\keyFont{\fontsize{8}{11}\helveticabold }
\def\firstAuthorLast{Jacques-Dumas {et~al.}} 
\def\Authors{Valérian Jacques-Dumas\,$^{1}$, Francesco Ragone\,$^{1,2,3}$, Pierre Borgnat\,$^{1}$, Patrice Abry\,$^{1}$, Freddy Bouchet\,$^{1,*}$}
\def\Address{$^{1}$Univ Lyon, Ens de Lyon, Univ Claude Bernard, CNRS, Laboratoire de Physique, Lyon, France \\
$^{2}$Earth and Life Institute, Université Catholique de Louvain, Louvain-la-neuve, BELGIUM \\
$^{3}$Royal Meteorological Institute, Brussels, BELGIUM }

\def\corrAuthor{Corresponding Author}

\def\corrEmail{Freddy.Bouchet@ens-lyon.fr}

\begin{document}
\onecolumn
\firstpage{1}

\title[Deep Learning Extreme Heatwave Forecast]{Deep Learning-based Extreme Heatwave Forecast} 

\author[\firstAuthorLast ]{\Authors} 
\address{} 
\correspondance{} 

\extraAuth{}

\maketitle

\begin{abstract}

Because of the impact of extreme heat waves and heat domes on society and biodiversity, their study is a key challenge. \FB{We specifically study long-lasting extreme heat waves, which are among the most important for climate impacts.} Physics driven weather forecast systems or climate models can be used to forecast their occurrence or predict their probability.  The present work explores the use of deep learning architectures, trained using outputs of a climate model, as an alternative strategy to forecast the occurrence of extreme \FB{long-lasting} heatwave. 
This new approach will be useful for several key scientific goals which include the study of climate model statistics, building a quantitative proxy for resampling rare events in climate models, study the impact of climate change, and should eventually be useful for forecasting. Fulfilling these important goals implies addressing issues such as class-size imbalance that is intrinsically associated with rare event prediction, assessing the potential benefits of transfer learning to address the nested nature of extreme events (naturally included in less extreme ones). We train a Convolutional Neural Network, using 1000 years of climate model outputs, with large-class undersampling and transfer learning. From the observed snapshots of the surface temperature and the 500 hPa geopotential height fields, the trained network achieves significant performance in forecasting the occurrence of long-lasting extreme heatwaves. We are able to predict them at three different levels of intensity, and as early as 15 days ahead of the start of the event (30 days ahead of the end of the event).  

\tiny
 \keyFont{ \section{Keywords:} Heatwave,
Extreme event, Atmosphere dynamics, Deep learning, Prediction.} 
\end{abstract}

\section{Introduction}
\label{sec:intro}

\noindent {\bf Context: Climate extreme event impacts and forecast.} Climate change constitutes one of the major concerns of modern societies. 
Its most severe impacts are caused by rare and extreme events. 
For instance, recent decades witnessed a number of exceptionally warm summers and record breaking heatwaves \cite{IPCC_2013}. 
At Northern Hemisphere mid-latitudes, relevant such examples were observed over Western Europe during the summer 2003 with a death toll of about $70,000$ \cite{garcia-herrera_review_2010}, or over Russia during the summer 2010 \cite{Barriopedro_2011,Otto_2012}, or over the North American Pacific coast \cite{philip2021rapid} during the summer 2021. \FB{The two main drivers of the death toll for the 2003 Western Europe heat wave, were the high level of temperature and the long duration (two successive heat events along an overall period of one month). The three main drivers of the impacts of the 2010 Russian heat wave, were the compound effect of high temperature, long duration (one month), and related fires. As this is key for impact, we study specifically in this work long-lasting extreme heat waves.} 

\FB{Those three extreme heat waves were unprecedented in historical data-sets.} Because of the scarcity of these events, estimating their return periods in preindustrial or current climates, estimating their occurrence probability, forecasting them several weeks in advance, or detecting early-precursors are major challenges in climate sciences.  The three main scientific difficulties stem from a lack of historical data for such unprecedented events, the difficulty to build reliable statistics using weather or climate models because of the huge numerical cost needed to obtain a sufficient number of events, and the quantitative assessment of model biases for those extremes given the scarcity of the data. Machine learning should be used in the future, in several different ways, in conjunction with physical models, to solve these three issues. In this paper we focus on a specific goal which is the forecast problem, which is a key starting point, for addressing several of these goals, as further discussed below.\\

\noindent {\bf Related work: study of long-lasting extreme heat waves and machine learning in climate sciences.} 
To define heatwaves, several indices have been used in the meteorology, climate, and impact literature, for different purposes \cite{perkins2015}. \FB{Quoting \cite{perkins2015}, "it seems that almost, if not every climatological study that looks at heatwaves uses a different metric". Many meteorology criteria and past climate studies of temperature extremes remained focused on intra-day data (see for instance \cite{IPCC_2013}). This might indeed be relevant for several application and risks. However, long-lasting heatwaves are the most detrimental to health \cite{Barriopedro_2011} and biodiversity.  Moreover, many of the extreme heatwaves with the largest impact, for instance the Western European one in 2003, the Russian one in 2010, or the North American Pacific coast one in 2021, lasted long, from two to five weeks. They were often composed of several sub-events with the classical definitions \cite{perkins2015}. The lack of comprehensive studies of the statistics of long-lasting events has actually been stressed in the last IPCC report \cite{IPCC_2021}. Moreover, many definitions that actually involve a measure related to the persistence of anomalous daily maximum temperature values with prescribed amplitude. Then they do not always carry a natural definition of a heatwave amplitude \cite{perkins2015}. This prevents to study independently impact of amplitude and duration of the heat-wave. This calls for a complementary definition of heatwaves, that can quantify both their amplitude in terms of temperature and their duration, in an independent way.}

\FB{Seminal studies \cite{schar2004role,Barriopedro_2011,coumou_decade_2012} of the 2003 and 2010 heatwaves already considered the averaged temperature over variable long time periods (7 days, 15 days, one month, three months). To deal with the goal of quantifying heat wave amplitudes for several independent duration, heatwave indices based on the combined temporal and spatial averages of the surface or 2-meter temperature has been adopted in a set of recent studies \cite{Ragone24, galfi2019, ragone2019, galfi_lucarini2021, ragone2021, galfi_ragone2021}.} This viewpoint is expected to be complementary with the classical definitions \cite{perkins2015}, and extremely relevant to events with the most severe impacts.  Moreover, such definitions have the advantage to define events which are spatially and temporally very precisely located, which is much better suited in a prediction and forecast perspective. Moreover the amplitude of a heatwave is naturally defined. In the present work, we follow this definition of long-lasting extreme heat waves, and assess their predictability using machine learning. 

Machine learning has now been used for decades in climate and weather forecast sciences with various goals, such as post-processing, data assimilation, physical analysis, etc.
Recently, deep neural networks were used with noticeable successes for prediction purpose \cite{weyn2019can,dueben2018challenges,scher2019weather}.
While deep learning-based prediction performance remains far from challenging the prediction capabilities of physics modelling-driven procedures \cite{weyn2019can}, they prove useful to improve physics models \cite{schneider2017earth} or their parameter tuning \cite{brenowitz2018prognostic,
gentine2018could}, to complementing them for analysis or pattern recognition \cite{liu2016application}, or to performing tasks not achievable with physics models. 
Deep learning has also been used for extreme weather event prediction \cite{pedram} or severe weather risk assessment \cite{mcgovern2017using}. In \cite{pedram}, it is shown that the CapsNet deep neural network is a fast and efficient tool, for predicting hot days several days ahead \FB{(intra-day heat waves)}. In \cite{LSTM}, it is shown that Long Short-Term Memory neural networks, focused this time on time series, are efficient in temperature prediction. \FB{As far as we now, no machine learning approach has been used so far to study long-lasting heatwaves. From a forecast point of view, compared to the prediction of intra-day heatwaves, this is a more difficult task as one should be able to perform a prediction over the sum of the number of days ahead of the heat wave and the heat wave duration. Moreover, the phenomenology is expected to be very different.}

\noindent {\bf Goals of this study.} \PA {This work intends to use} machine learning to predict the future occurrence of long-lasting extreme heat waves. As far as we now, this is the first use of machine learning for this goal. The learning uses 1000 years of outputs of a climate model. From this data, our algorithm predict, from the observation of the surface temperature and 500 hPa geopotential height, whether a long-lasting heat wave that starts within $\tau$ days, will occur. We focus on this prediction task in this article, using climate model data. We argue in the conclusion how this machine learning prediction algorithm should be useful in the future as a key element to solve the three major scientific challenges of climate extreme studies: lack of historical data, the issue of model sampling, and model bias studies.  
\PA{More precisely, the present work targets specific goals:} 
first advancing the machine learning methodology to study rare long-lasting heatwaves~; 
second \PA{performing} the first study of the predictability of those extreme events.

The first specific goal is to evaluate deep learning-based architectures for the heatwave forecast problem, and to quantify their \PA{performance}, \PA{so as to avoid the recourse to an arbitrary choice of features as commonly done in classical machine learning}. To that end, we build and train a classifier from a data-set of outputs of the planet simulator (PLASIM) climate model. Such data can be made available in reasonable size, and at reasonable costs. \modPB{It allows us to formulate the problem as supervised classification, and to train a deep learning method}. 

We address a number of crucial methodological issues: 
i) propose a suitable deep learning architecture; 
ii) overcome the imbalance in class-sizes intrinsically associated with extreme events and requiring the use of sampling strategies; and
iii) account for the nested nature of extreme events (most extreme events are included in less extreme ones). This last point suggests the potential use of transfer learning and we will study this aspect.

Defining long-lasting heat waves as temporal and spatial averages, the first works \cite{Ragone24, galfi2019, ragone2019, galfi_lucarini2021, ragone2021, galfi_ragone2021} focused, on the one hand, on discussing their statistics and probability, and on the other hand, on improving the statistics of extremely rare events using rare event algorithms \cite{Ragone24, ragone2019,  ragone2021}. The second goal of this work will be to assess the predictability and forecast potential of these extreme heat waves. We will achieve it by showing that the trained network can indeed predict long-lasting heat waves up to 15 days ahead of the start of the event.





\noindent {\bf Contributions and outline.} 
The climate model, the dataset of model outputs, and the definition of long-lasting heat waves are discussed in Section~\ref{sec-data}.  The machine learning methodology, dealing with class imbalance, and transfer learning are addressed and discussed in Section~\ref{sec:dl}.

Results are reported in Section~\ref{sec:results}. We first compare aggregation protocols aiming to best combine different available observations, and second discuss the benefits of using transfer learning in nested extreme events prediction strategies as well as non-extreme event large class undersampling, in Section~\ref{sec:resultsa}. Furthermore, the significant ability of Convolutional Neural Network-based deep learning architectures to perform relevant forecast of the occurrence of long-lasting extreme heatwaves, several days in advance is quantified in Section~\ref{sec:resultsb}. Notably, it is shown that the occurrence of heatwaves can be predicted up to $\tau = 15$ days in advance, thus significantly beyond typical correlations times for climate data of the order of 3 to 5 days  \cite{vallis2017atmospheric}. 

Finally we discuss in Section~\ref{sec:conclusion} perspectives for using the deep learning-based forecast of extreme heatwaves, as a key element to tackle the three key scientific challenges of climate extreme events.  

\section{Data and Methods}

\subsection{Climate data and heatwaves}
\label{sec-data}

\noindent {\bf Climate model data.} 
\modPB{As explained in the Introduction,  because of the lack of historical data for unprecedented heatwaves, data-based heatwave forecast must necessarily start from model data. Hence, we use simulated climate model outputs as a training set for the task of classifying whether a given observation of the atmosphere leads to extreme events. We also reserve a part of the simulation to test the prediction and compute its accuracy.}

The data used in the present work are produced by the Planet Simulator (PlaSim) climate model \cite{Plasim}, \cite{puma}, as computed for the work \cite{Ragone24}.
Its dynamical core solves the primitive equations for vorticity, divergence, temperature and surface pressure. 
Moisture is included by transport of water vapor. 
The governing equations are solved using a spectral transform method. 
Unresolved processes, such as radiation, interactive clouds, moist and dry convection, large-scale precipitation, boundary layer fluxes of latent and sensible heat and vertical and horizontal diffusion are parametrized. 
The model also simulates the coupling with land surface scheme and ocean. 

The horizontal resolution is  T$42$ in spectral space, corresponding to a spatial resolution of about $2.8$ degrees in both latitude and longitude. In practice, the horizontal fields of data have a spatial size of $64\times128$ pixels, covering the entire globe.
The vertical resolution corresponds to $10$ vertical layers. 
\PA{Moreover, each field is sampled in time at $\delta t = 3$hours sampling period.} 

The model is setup to run with fixed greenhouse gases concentrations and boundary conditions (incoming solar radiation, sea surface temperature and sea ice cover distributions) cyclically repeated every year, in order to generate a stationary state reproducing a climate close to the one of the $1990$'s. The simulation has been run so that a thousand of physical years of model outputs are available. They were computed on $16$ processors and the total simulation took $1111$ hours to compute, hence with a moderate cost.\\

 \begin{figure*}
     \centering
     \includegraphics[width=\linewidth]{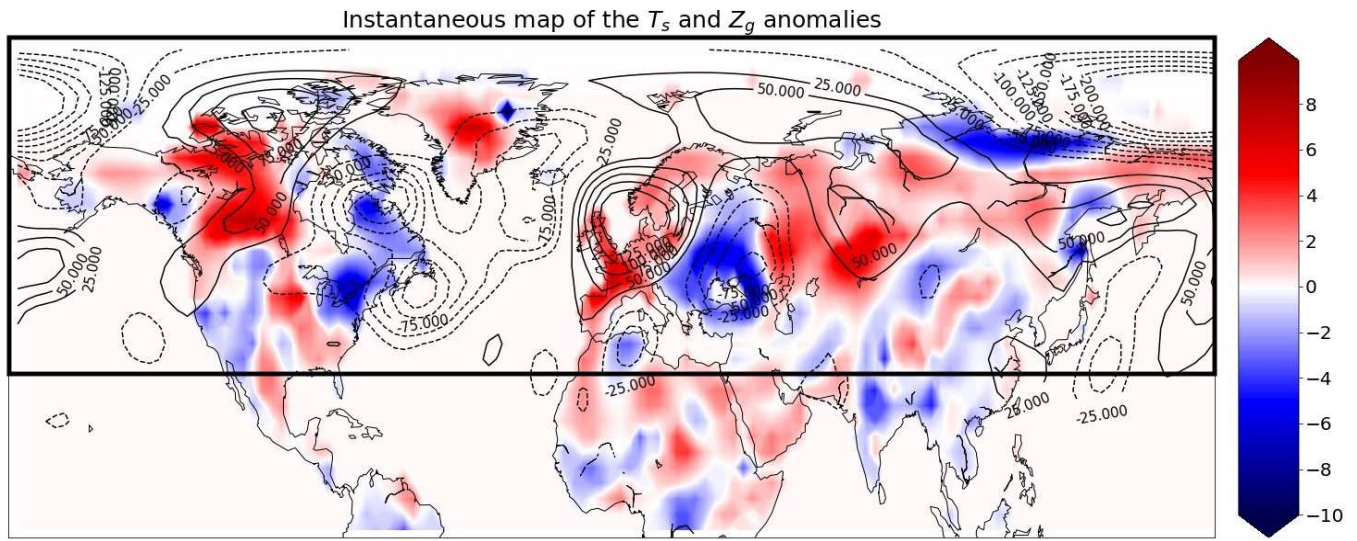}
     \caption{Snapshot of the surface temperature surface fluctuations ($T_s$ fluctuations, according to the color bar, in Kelvin) and of the geopotential height at $500$mbar ($Z_g$ fluctuations, contours) over the Northern Hemisphere. This snapshot is taken on July $20^{th}$ on a arbitrary year of the PlaSim simulation. The spatial resolution is $64\times128$ (latitude$\times$longitude). 
     The thin contour lines, representing the anomaly of $Z_g$, are in meters. The thick black contour delimits the zone that is used for prediction by the machine learning procedure.}
     \label{fig:original_map}
 \end{figure*}

\noindent {\bf Climate data and heatwaves.} 
The present work focuses on predicting summer heatwaves. 
For that, two horizontal fields classically associated with heatwave mechanisms \cite{Ragone24} are used only amongst the very large size PlaSim outputs: the surface temperature $T_s$ (in Kelvin) and the height $Z_g$ (in meters) of the geopotential on the isopressure surface of $500$ hectoPascal (hPa), located in the middle troposphere. 
The relation between surface temperature and heatwaves is straightforward.
In weather and climate dynamics, the geopotential height in the middle of the troposphere is considered an excellent representation of the dynamical state of the atmosphere.
Indeed, the geopotential height (in meters) at 500-hPa, $Z_g$ is further tightly related to anticyclones (positive values) and cyclones (negative values) in the lower atmosphere. 
Moreover, to a good approximation, the wind flows along the isolines of the geopotential height.\\

\noindent {\bf Heatwave definition.} 
Let us precisely define heatwaves, as proposed for the present work and following recent studies \cite{Ragone24, galfi2019, ragone2019, galfi_lucarini2021, ragone2021, galfi_ragone2021}. For that, it is first needed to define the fluctuations in temperature and geopotential height, which are called anomalies when the fluctuations are large.
Let $T_s\left(\vec{r},t\right)$ denote the surface temperature at location $\vec{r}$ and time index $t$, where time is counted independently from $0$ for each year and sampled at a 3-hour resolution.
The ensemble average $\left<T_s\right>(\vec{r},t)$ is obtained as the average across the $1000$ years of $T_s(\vec{r},t)$ for each given location $\vec{r}$ and intra-year time $t$, thus preserving intra-year seasonal effect.
The temperature fluctuation is further defined as $\left(T_s-\left<T_s\right>\right)(\vec{r},t)$.
Geopotential height fluctuation is defined accordingly.
A snapshot of maps of temperature and 500-hPa geopotential height fluctuations is shown in Fig.~\ref{fig:original_map}.

We define $Y(t)$ the time-space average of the temperature fluctuations as
$$Y(t) = \displaystyle{\frac{1}{D}\int_{t}^{t+D}\frac{1}{\left|\mathcal{A}\right|}\int_\mathcal{A}\left(T_s-\left<T_s\right>\right)(\vec{r},u)\mathrm{d}\vec{r}\mathrm{d}u} $$
over the region $\mathcal{A}$ and a duration $D$, at time $t$. $\left| \mathcal{A} \right|$ is the area of the region $\mathcal{A}$. A heatwave of duration $D$ and of strength $a$, is said to occur at time $t$ when $Y(t)>a$.
For the present work, we study summer heatwaves (occurring in June, July, August only) over France ($\equiv \mathcal{A}$) lasting for $D=14$ days.

By nature, extreme heatwaves constitute rare events. We will consider three strength levels for $a$, defined as the $5\%$, $2.5\%$ or $1.25\%$ most extreme events. From the data, it gives thresholds in time-space average of temperature fluctuations of $a_5 =3.08$K, $ a_{2.5}=3.7$K, and $a_{1.25}=4.23$K respectively. 

\FB{As explained in the introduction, this definition of heat-waves follows seminal studies \cite{schar2004role,Barriopedro_2011,coumou_decade_2012} of the 2003 and 2010 heatwaves; it is specifically suited for the study of high impact events, and has been adopted in a set of recent studies \cite{Ragone24, galfi2019, ragone2019, galfi_lucarini2021, ragone2021, galfi_ragone2021}.}
\\


\noindent {\bf Heatwave prediction dataset.} For the prediction of heatwaves over France, data are restricted to dynamically relevant areas: North Hemisphere mid-latitudes, above $30^\circ$N. On Fig.~\ref{fig:original_map}, it corresponds to the thick black box, and the size of the fields is then $25\times128$ pixels at the model resolution. 

Instead of the direct use of data in the physical space, which would imply to handle spherical geometry and related boundary conditions, it has been chosen here to work with their spatial Fourier Transform (FT), computed on a $64\times64 $ grid, with a frequency resolution of approximately $\delta F \simeq 10^{-4}\text{km}^{-1}$ in each direction.

The data used as inputs of the supervised learning procedure described in Section~\ref{sec:dl} below thus consist of couples ($X_t$, $Z_t$), for  $t$  ranging from June 1st to August 31st, for $1000$ years of simulation.
Vector $Z_t$ denotes a binary label, with value $1$ when $Y(t)>a$, i.e., when there is an occurrence of heatwave in the next D-days, and $0$ otherwise. $X_t$ stands for the $64\times64\times2$ spatial FT $\tilde{T}_s$ and $ \tilde{Z}_g$ of fields $T_s(t-\tau)$, and $Z_g(t-\tau)$. $\tau$ here denotes the delay (in days) between the date of observations and the date at which a prediction of heatwave occurrence is to be made. If $\delta t$ is the time lag between two consecutive samples (with $\delta t=3$h in PlaSim), then we have $\tau=8\times\delta t$.
In other words, to make a prediction of heatwave occurring sometimes between today and the next $D$ days, data observed $\tau$ days prior to today are used.

\subsection{Deep learning architecture and procedure}
\label{sec:dl}

\begin{figure}[ht]
\centerline{\includegraphics[width=\columnwidth]{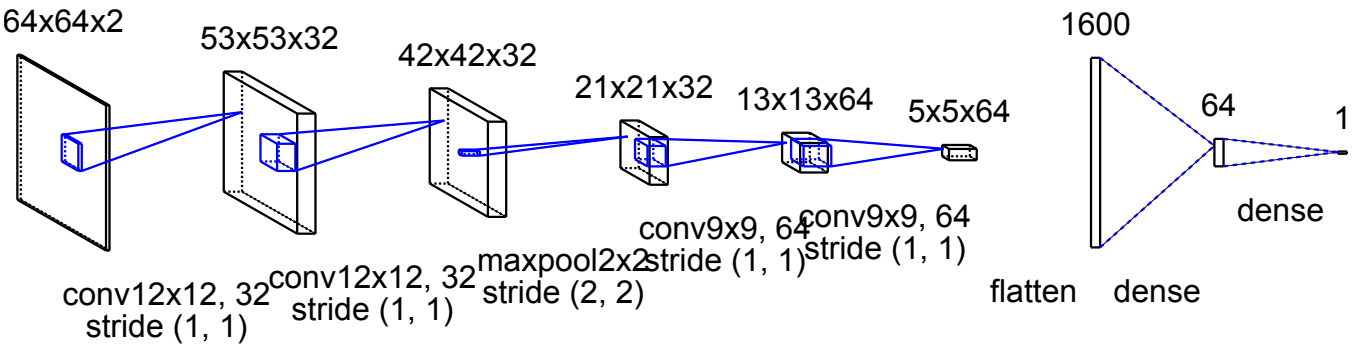}}   
     \caption{\label{fig:cnn} {\bf CNN-based heatwave predictor architecture.}
     \modPB{For stacked data (see aggregation protocol P4), the first layer has a size of $64 \times 64 \times  4$.}}
\end{figure}

\noindent {\bf Convolutional Neural Network (CNN) architecture.} Heatwave prediction is performed as a supervised classification problem, using the CNN-based deep-learning $4$ layer-architecture depicted in Fig.~\ref{fig:cnn}. 
\modPB{The choice of CNN instead of classical ML methods stems from the high dimension of the data: other methods would require to engineer arbitrary features.}

In the proposed architecture, the first two convolutional layers have filters of size $12 \times 12$ and ReLU activation functions. 
They are followed by a maxpool layer so as to divide data size by $2 \times 2$ with spatial dropout.
The next two convolutional layers have filters of size $9 \times 9$ and ReLU activation functions and are also followed by a spatial dropout.
Finally, a flatten operation and two fully connected layers followed by a sigmoid activation yield an output between $0 \leq q \leq 1$. This output is associated with the probability of occurrence of an heatwave in the upcoming $D$ days.

\noindent {\bf Train/test sets.} Typical time scales governing climate dynamics permit to consider the $1000$ year PlaSim simulation data as $1000$ independent trajectories. Yet, there are significant intra-year spatiotemporal dependencies, that can be exploited for heatwave detection.
Therefore, the training set does not select events at random in time index and uniformly across the entire dataset.
Instead, the split between train/test set is based on the random sampling of full years, with $900$ such years associated with the train set, while the test set comprises of the $100$ remaining years. 
The overall training set thus gathers $K=648000$ samples ($900\ \text{years}\times3\ \text{months}\times30\ \text{days}\times8$ samples per day). 
\modPB{Furthermore, the prediction will be about extreme events. We need to be certain that the test set will contain enough of these extreme events so that the evaluation is fair. Hence, we constrain the test set to have the same proportion of extreme events as in the training set. This prevents imbalance of events for the evaluation; yet, this does not solve yet the issue of imbalanced data (in each train/test set, data are imbalanced).}
\\

\noindent {\bf Learning parameters.} Learning architectures, training and testing are implemented using Python with Keras API. 
For optimization, the AMSGrad variant~\cite{DBLP:conf/iclr/ReddiKK18} is used, with a learning rate of $2.10^{-4}$ and momentum of $0.5$. 
The dropout rate is set to $30\%$. 
Batch size is set to $1000$ samples. 
Batch normalization \cite{batch_normalization} is applied after each layer.
The number of training epochs\footnote{The number of epochs is the number of training iterations over the whole dataset, here the 900 years of the train set.} is set  empirically to $10$ when using the threshold $a_5$, and to $5$ for the two other thresholds $a_{2.5}$ and $a_{1.25}$.
As the problem consists in detecting the absence or presence of heatwave (based on temperature anomaly), the loss function is the standard Binary Cross-Entropy, commonly used for supervised classification tasks \cite{Goodfellow-et-al-2016}. \\

\noindent {\bf Class-size imbalance and undersampling.} 
For the prediction of rare events, classes are imbalanced by construction. It has been well-documented that machine learning training is severely impaired by imbalanced class sizes \cite{imbalance,Johnson-Khoshgoftaar-2020}.
Here, we propose to handle this by \emph{undersampling} \modPB{the training set (only)}: only a fraction $S_a$ of (randomly selected) non-heatwave samples are used.
A natural starting idea is to ensure, on average, equal sizes for both classes. The class $a_5$, for instance, contains $5\%$ of positive event, the dataset thus contains $19$ times as many negative events (since $20\times5\%=100\%$). This leads to subsampling the non-heatwave class by factors $S_a$ of $1/19$, $1/39$ and $1/79$, respectively for the three heatwave levels $a_5, a_{2.5}$ and $a_{1.25}$. 
Less severe subsampling rates are also tested by considering multiplying by $s$ the previous subsampling rates, with $s=2, s=4$ and $s=10$. For instance, when $s=2$, we considered for simplicity that the subsampling factors were $1/10$, $1/20$ and $1/40$ (instead of $2/19$, $2/39$ and $2/79$), respectively for the three heatwave levels $a_5, a_{2.5}$ and $a_{1.25}$. It means that in this case, each dataset contains approximately twice as many negative events as positive events, and so on for the different values of $s$. We also compared this random undersampling with the case where no undersampling was applied.
This subsampling procedure yields training set of size $K_{a,s} = K \times p_a + K \times (1-p_a) \times S_a \times s $, where $p_a = 0.05, 0.025, 0.0125$ corresponds to the fraction of most extreme events associated with threshold $a_5,a_{2.5}$ and $a_{1.25}$. \modPB{For clarification, note that the test set is used without any undersampling (knowledge of the existence of heatwave is only used to compute performance).}

For the larger heatwave levels, undersampling reduces a lot the number of training samples and will degrade the performance. In this situation, we rely on a second technique: warm-start transfer learning so as to leverage the larger size of the available training set at lower level heatwaves.
\\

\noindent {\bf Transfer learning.} Heatwave detection is performed for three different intensity levels. 
The direct approach is to train the learning  procedure for each of the three levels, independently and using random initialization. The weights are initialized using the Glorot uniform initializer \cite{glorot}.

However, as we consider increasingly extreme heatwaves, the datasets contain fewer and fewer positive events, which makes the learning task increasingly harder for the neural network when starting from scratch. A more elaborated second approach is then proposed, using transfer learning \cite{transfer_learning} for the two highest levels. The idea is to ease the learning task by using information previously learnt with less extreme heatwaves. It consists in three steps: 
\begin{itemize}
    \item (i) Learning is first performed from scratch for the $5\%$ most extreme events
    \item (ii) Learning for the top $2.5\%$ is performed, being initialized with the weights of the CNN learnt for the $5\%$ heatwave level; this is a warm-start transfer learning. 
    \item (iii) Learning for the top $1.25\%$ is performed, being initialized with the weights of the CNN learnt for the $2.5\%$ heatwave level
\end{itemize}

Of importance, to ensure meaningful statistical performance assessment, the same train/test sets are used during learning for the three levels of heatwaves, both with and without transfer learning.\\

\noindent {\bf Performance assessment.} To assess quantitatively the relevance of the proposed CNN-based heatwave occurrence prediction, the train/test procedure for the three levels of heatwaves, with and without transfer learning, is repeated $40$ times, with independent train/test data split, respecting the procedure described above. 
For each of these 40 trials, detection performance is assessed by computing, on the test set, \modPB{the rates (in percentage) of True Positives (TPR), True Negatives (TNR), False Positives (FPR) and False Negatives (FNR). As usual, TPR are computed as: 
$$\mathrm{TPR} = \frac{\mathrm{Number\ of\ correctly\ predicted\ heatwaves}}{\mathrm{Number\ of\ actual\ heatwave\ events}}.$$

FPR are defined accordingly as: $$\mathrm{FPR} = \frac{\mathrm{Number\ of\ FP (wrongly\ predicted\ heatwaves)}}{\mathrm{Number\ of\ negative\ events\ (no\ heatwave)}}.$$ As TPR+FNR=1 and FPR+TNR=1, we will report and comment only TPR and FPR.} Then, the Matthews Correlation Coefficient (MCC) \cite{MCC} is reported, that is a single number score that balances the Type-I (FP) vs. Type-II (FN) errors (false alarm for a heatwave vs. non detection when one occurs); a main advantage is that it accounts for imbalance between class sizes while allowing to compare performance in different situations with only one number. The values of MCC presented in the Tables and Figures are all computed on test sets. Means and standard deviations (and maximum absolute deviations in Fig.~\ref{fig:compare_vs_tau}) of these scores are computed by average across trials. 

\PA{\noindent {\bf Robustness and reproductibility.} To assess the robustness and reproductibility of the prediction performance reported in Section~\ref{sec:results}  with respect to the chosen architecture, we have systematically applied a repeated learning from scratch procedure: it consists in performing 40 times independently the training of the network and the quantification of performance, using different initializations and independent train/test splits. 
Prediction performance are systematically given as mean, standard deviations, best and worst cases. 
To assess the impact of the architecture details, a number of different CNN architectures were tested, varying the number of layers, the size of filters, the parameters of the Dropout and MaxPool layers, the size and number of Dense layers, the reduction of the size of the data. 
Reporting results for each architecture would have resulted in a lengthy paper.
Our main conclusion is that prediction performance are essentially similar across a large range of variations of parameters. 
Performance are reported for the architecture detailed in Fig.~\ref{fig:cnn} that correspond to typical performance reported in the (large) subset of architecture yielding equivalent the best performance. 
}

\begin{table}[t]
    \centering
    \begin{tabular}{c||c|c|c||c|c|c}
    Without transfer learning & \multicolumn{3}{c||}{TPR in \%: average (std)} & \multicolumn{3}{c}{FPR in \%: average (std)} \\
    \hline
    Levels & $5\%$ & $2.5\%$ & $1.25\%$ & $5\%$ & $2.5\%$ & $1.25\%$\\
    \hline
    FT of $T_s$ alone (P1) & 43.4 (11.1) & \bf 39.7 (16.7) & \bf 33.6 (22.7) & 5.2 (2.9) & 3.0 (2.5) & 2.3 (2.7)\\
    FT of $Z_g$ alone (P2) & \bf 56.3 (12.9) & 38.3 (14.5) & 22.7 (14.6) & 14.2 (6.0) & 10.3 (5.1) & 7.5 (5.9)\\ 
    Events in common & 29.9 (8.3) & 18.7 (10.4) & 7.8 (8.1) & \bf 2.1 (1.1) & \bf 0.9 (0.8) & \bf 0.3 (0.4)\\
    \hline
    With transfer learning & \multicolumn{3}{c||}{True Positives Rates} & \multicolumn{3}{c}{False Positives Rates}\\
    \hline
    Levels & $5\%$ & $2.5\%$ & $1.25\%$ & $5\%$ & $2.5\%$ & $1.25\%$ \\
    \hline
    FT of $T_s$ alone & $-$ & 30.4 (9.0) & 24.6 (9.4) & $-$ & 2.0 (0.8) & 1.1 (0.5) \\
    FT of $Z_g$ alone & $-$ & \bf 46.4 (14.2) & \bf 38.4 (12.8) & $-$ & 8.1 (3.9) & 5.1 (2.7) \\ 
    Events in common & $-$ & 19.4 (7.6) & 13.5 (5.9) & $-$ & \bf 0.7 (0.3) & \bf 0.3 (0.2) \\
    \end{tabular}
    
    \begin{tabular}{c||c|c|c}
    Without transfer learning & \multicolumn{3}{c}{MCC: average (std)}\\
    \hline
    Levels & $5\%$ & $2.5\%$ & $1.25\%$\\
    \hline
    FT of $T_s$ alone & 0.33 (0.04) & 0.30 (0.06) & 0.23 (0.09) \\
    FT of $Z_g$ alone & 0.25 (0.03) & 0.14 (0.04) & 0.06 (0.03) \\ 
    Events in common & \bf 0.44 (0.06) & \bf 0.33 (0.09) & \bf 0.15 (0.12) \\
    \hline
    With transfer learning & \multicolumn{3}{c}{Average MCC}\\
    \hline
    Levels & $5\%$ & $2.5\%$ & $1.25\%$ \\
    \hline
    FT of $T_s$ alone & $-$ & 0.27 (0.05) & 0.23 (0.07) \\
    FT of $Z_g$ alone & $-$ & 0.21 (0.03) & 0.17 (0.03) \\ 
    Events in common & $-$ & \bf 0.35 (0.09) & \bf 0.28 (0.09) \\
    \end{tabular}
    \caption{\label{tab:confusion_compare}
    \modPB{ {\bf Compared performance for heatwave occurrence prediction from surface temperature vs. geopotential height.} We report in percentage the True positive/False positive Rates (average, standard deviations are in parenthesis) for prediction, for each heatwave levels ($5\%, 2.5\%$ and $1.25\%)$, with and without transfer learning. The last columns are the MCC: average value (and standard deviations) calculated across $40$ independent learning. 
    The first two lines correspond to $T_s$ alone and $Z_g$ alone, while the last line quantifies a logical \emph{AND} (i.e., joint prediction by $T_s$ and $Z_g$). Percentages of True Positives (resp. False Positives) rates are quantified with respect to the sizes of the positive (resp. negative) class. Results in bold shows the best result for each category (levels ; no or with transfer learning).}
    }
\end{table}

\begin{table*}[t]
    \centering
    \begin{tabular}{c||c|c|c||c|c|c}
    Without transfer learning & \multicolumn{3}{c||}{TPR in \%: average (std)} & \multicolumn{3}{c}{FPR in \%: average (std)}\\
    \hline
    Levels & $5\%$ & $2.5\%$ & $1.25\%$ & $5\%$ & $2.5\%$ & $1.25\%$\\
    \hline
    Events in common & 29.9 (8.3) & 18.7 (10.4) & 7.8 (8.1) & \bf 2.1 (1.1) & \bf 0.9 (0.8) & \bf 0.3 (0.4)\\
    FT of $T_s$ and FT of $Z_g$ (P3) & 44.6 (7.0) & 21.4 (5.7) & 8.0 (4.1) & 4.7 (1.6) & 3.7 (1.6) & 1.7 (1.4)\\
    Stacked FT of $T_s$ and $Z_g$ (P4) & \bf 69.8 (8.5) & \bf 37.9 (4.0) & \bf 14.8 (4.0) & 7.6 (2.6) & 7.7 (2.2) & 4.7 (2.9)\\ 
    \hline
    With transfer learning & \multicolumn{3}{c||}{True Positives Rates} & \multicolumn{3}{c}{False Positives Rates}\\
    \hline
    Levels & $5\%$ & $2.5\%$ & $1.25\%$ & $5\%$ & $2.5\%$ & $1.25\%$\\
    \hline
    Events in common & $-$ & 19.4 (7.6) & 13.5 (5.9) & $-$ & \bf 0.7 (0.3) & \bf 0.3 (0.2) \\
    FT of $T_s$ and FT of $Z_g$ & $-$ & 35.2 (8.2) & 27.7 (10.6) & $-$ & 2.4 (0.7) & 1.2 (0.7) \\
    Stacked FT of $T_s$ and $Z_g$ & $-$ & \bf 64.2 (9.8) & \bf 58.5 (13.2) & $-$ & 4.4 (1.5) & 2.7 (1.1) \\
    \end{tabular}
    
    \begin{tabular}{c||c|c|c}
    Without transfer learning & \multicolumn{3}{c}{MCC: average (std)}\\
    \hline
    Levels & $5\%$ & $2.5\%$ & $1.25\%$\\
    \hline
    Events in common & \bf 0.44 (0.06) & 0.33 (0.09) & 0.15 (0.12) \\
    FT of $T_s$ and FT of $Z_g$ & 0.35 (0.03) & 0.29 (0.05) & 0.25 (0.08) \\
    Stacked FT of $T_s$ and $Z_g$ & \bf 0.44 (0.03) & \bf 0.37 (0.03) & \bf 0.29 (0.06) \\ 
    \hline
    With transfer learning & \multicolumn{3}{c}{Average MCC}\\
    \hline
    Levels & $5\%$ & $2.5\%$ & $1.25\%$\\
    \hline
    Events in common & $-$ & 0.35 (0.09) & 0.28 (0.09) \\
    FT of $T_s$ and FT of $Z_g$ & $-$ & 0.29 (0.04) & 0.24 (0.07) \\
    Stacked FT of $T_s$ and $Z_g$ & $-$ & \bf 0.40 (0.04) & \bf 0.35 (0.05) \\
    \end{tabular}
    \caption{\label{tab:scores}
    \modPB{ {\bf Compared performance for heatwave occurrence prediction with different data aggregation protocols.} We report in percentage the True positive/False positive Rates (average, standard deviations are in parenthesis) for prediction, for each heatwave levels ($5\%, 2.5\%$ and $1.25\%)$, with and without transfer learning. The last columns are the MCC: average value (and standard deviations) calculated across $40$ independent learning (as in Table \ref{tab:scores}). 
    The first line is the same logical \emph{AND} as in Table \ref{tab:confusion_compare} for reference. The next two lines correspond to two ways of combining the FT of $T_s$ and $Z_g$: a simple combination of information independently learnt on the second line and the FT stacked on the third line. Percentages of True Positives (resp. False Positives) rates are quantified with respect to the sizes of the positive (resp. negative) class. Results in bold shows the best result for each category (levels ; no or with transfer learning).}
    }
\end{table*}

\section{Results for extreme heatwaves prediction}
\label{sec:results}

This section will report the results obtained while using all or only parts of the proposed methodology.

\subsection{Data aggregation, undersampling and transfer learning}
\label{sec:resultsa}

To address the methodological issues of data aggregation, training set undersampling and transfer learning, analyses first concentrate on the easiest case $\tau= 0$ ($\tau$ being the delay in days between the date of the prediction and the start date of the heatwave).
Let us emphasize however that predicting the occurrence of heatwave at $\tau = 0$ is already far from a trivial endeavour, as it implies predicting from data at time $t$, the existence of heatwaves occurring at any time between $t$ and $t+D$ ($D$ being in days the duration of the heatwave). \\

\noindent {\bf Surface temperature versus geopotential height.} As described in Section~\ref{sec-data}, data available for heatwave predictions consist of the $64\times64\times2$ spatial FT $\tilde{T}_s(t) $ and $\tilde{Z}_g(t) $ of $T_s(t)$ and/or $Z_g(t)$, respectively, for each time position $t$. Table~\ref{tab:confusion_compare} first compares forecasting performance from two independent learning protocols: \\
\indent P1) $T_s$-only,  using $\tilde{T}_s(t) $ alone as a $64\times64\times2$ tensor CNN input~; \\
\indent P2) $Z_g$-only, using $\tilde{Z}_g(t) $ alone as a $64\times64\times2$ tensor CNN input.\\
\indent Table~\ref{tab:confusion_compare} shows first that surface temperature and geopotential height independently contain enough spatial structures to predict heatwave occurrences, even for the most extreme events, with MCC that positively departs from $0$. 
Table~\ref{tab:confusion_compare} however also clearly shows that surface temperature as input alone outperforms geopotential height as input alone in terms of MCC, which is especially true for the most extreme events.
Interestingly, Table~\ref{tab:confusion_compare} further shows that the poorest performance of geopotential height comes from much higher rates of False Positives. 
This may come as no surprise since heatwaves are intrinsically defined in terms of surface temperature fluctuations. 

Then, we probe whether the detected events by using $T_s$ or $Z_g$ are the same or not. This is the line 'Events in common' in Table~\ref{tab:confusion_compare}; it shows that FPR of events common between (independent) predictions from surface temperature and geopotential height is low. This suggests to combine these two independent detections to take advantage of the joint information available in these two fields. 
A naive and straightforward approach consists in performing a logical \emph{AND} between the outputs of the two independent predictions. 
\modPB{Table~\ref{tab:confusion_compare} indicates that the resulting MCC increases slightly (especially with transfer learning for events at 1.25\%). Yet, this comes with the price of a big reduction of TPR as the method only predicts True Positive events if it can be predicted from both fields ($T_s$ and $Z_g$); the MCC is good because the FPR is really small. This calls for more advanced data aggregation procedures, where the training can be done end-to-end using both fields at the same time, and hoping to obtain both good TPR and MCC. }\\ 

 \begin{figure*}
     \centering
     \includegraphics[width=\linewidth]{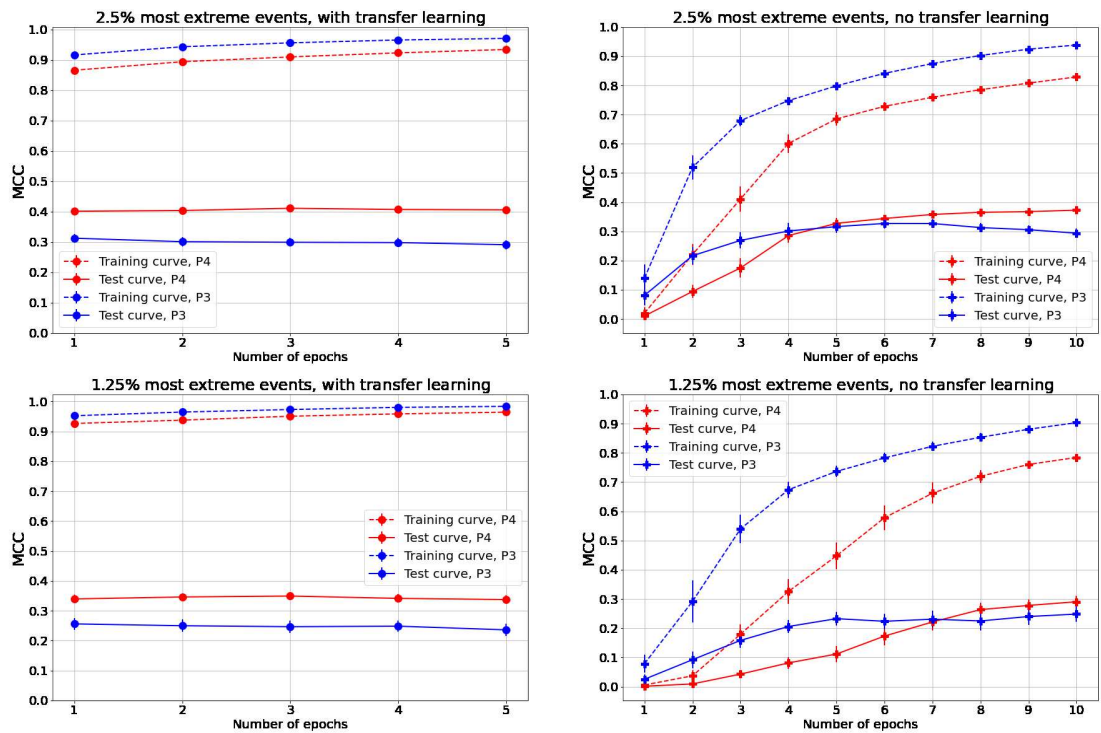}
     \caption{\label{fig:compare} 
     {\bf Forecasting performance in terms of MCC as a function of the number of epochs,} for aggregation protocols P3 (Combined-$T_sZ_g$) (blue) and P4 (Stacked-$T_sZ_g$) (red), without ('+', plots on the right) and with ('o', plots on the left) transfer learning. Solid (resp., dashed) lines correspond to testing (resp. training) performance. Top and bottom plots correspond respectively to the $2.5\%$ and $1.25\%$ most extreme events. Average MCC are obtained from $40$ independent learning. \modPB{Note that the discrepancy of values of MCC between training and test sets is due to the difference in class imbalance: training (solid lines) set is undersampled and the reduction in class imbalance (here: only 2) leads to a large MCC, while test set (dashed lines) has a high class imbalance ratio by definition of the problem of predicting extreme events and cannot achieve the same MCC. 
     }}
 \end{figure*}

\noindent {\bf Data aggregation.} 
Two new learning protocols based on aggregation of surface temperature and geopotential height data are proposed here. We defined them as:\\
\indent  P3) Combined-$T_sZ_g$, using both $\tilde{T}_s(t) $ and $\tilde{Z}_g(t) $, while using each of them as a $64\times64\times2$ tensor input of an independent CNN with same architecture as that described in Section~\ref{sec:dl}, but for the last fully-connected layer: the flattened outputs of both CNN are concatenated to serve as the input of a single final fully-connected layer; \\
\indent  P4) Stacked-$T_sZ_g$, using jointly $\tilde{T}_s(t) $ and $\tilde{Z}_g(t) $ by stacking them into a $64\times64\times4$ tensor used as the input of the CNN. \modPB{Let us point out that from the first layer, data  $\tilde{T}_s(t)$ and $\tilde{Z}_g(t)$ are then combined thanks to the summation to obtain one map.}\\
\indent Table~\ref{tab:scores} reports the forecasting performance of these four protocols in terms of TPR, FPR and MCC (averages and standard deviations obtained from $40$ independent learning). We checked that the median of the MCC is systematically close, with differences lower than $0.02$, to the mean MCC.
Comparing Table~\ref{tab:confusion_compare} and Table~\ref{tab:scores} first strikingly shows that aggregation protocol P3 (Combined-$T_sZ_g$) does not outperform the much simpler and less costly logical \emph{AND} based combination of protocols P1 ($T_s$-only) and P2 ($Z_g$-only). Although P3 improves the TPR, which is better for our prediction task, it also detects a much larger proportion of False Positives. This Table also shows that protocol P4 outperforms all the others and this is particularly clear with transfer learning. This method is the one yielding the largest proportion of False Positives, but the MCC is high thanks to the very high number of True Positives events predicted: still more than $50\%$ for the $1.25\%$ most extreme events.
This clearly indicates that the cross-spatial dynamics of surface temperature and geopotential height also contains relevant information pertaining to heatwave production mechanisms.
This further suggests that these cross-spatial dynamics are better exploited and revealed when the fields mixed and combined together from the first layer of the deep learning architecture layer (and thus from finest available spatial dynamics scales), as in aggregation protocol P4 (Stacked-$T_sZ_g$), rather than when processed independently and combined at the last (decision, and coarse scale) layer, as in aggregation protocol P3 (Combined-$T_sZ_g$).\\

\noindent {\bf Transfer learning.} To quantify the benefits of using transfer learning, prediction performance are compared when the training is performed independently for the three anomaly levels (hence without transfer learning) against when the training is performed with initialization of the weights of the network for a given heatwave level using the network weights learned from the immediately lower heatwave level. The weights of the training for the $2.5\%$ heatwave level are initialized with the weights learned at the $5\%$ heatwave level (i.e., the weights learned after $10$ epochs). In the same way, the weights of the training for the $1.25\%$ heatwave level are initialized with the weights obtained at the end of the training for the $2.5\%$ heatwave level (i.e., after $5$ epochs).
Prediction performance achieved in terms of MCC (averaged over the $40$ independent learning), with the four protocols, with and without transfer learning, are compared in Table \ref{tab:scores}. The runs performed without transfer learning systematically consisted in $10$ epochs and the MCC presented in Table \ref{tab:scores} is the average of the best MCC obtained among these $10$ epochs for each run. 
\\
\modPB{The results indicate that heatwave prediction performance, in terms of increased MCC, achieved with transfer learning is consistently comparable to without transfer learning; it is slightly above with stacking protocol P4 (having best performance).
In the case of protocols P1 and P2, we see also a small reduction of the standard deviations (computed across independent trials) when using transfer learning, thus indicating a weaker sensitivity to weight initialization prior to training. It is not so clear with protocols P3 and P4: on the one hand, the transfer learning leads to a decrease of the FPR associated with a decrease of the standard deviations in the FPR. On the other hand, the clear increase of TPR with protocol P4 goes along an increase of the TPR standard deviation, especially for the least extreme class of heatwaves.}
A main advantage is the reduction of the number of epochs required to train the system for rare events. This last point has been explored. For that, Fig.~\ref{fig:compare} reports the training performance with and without transfer learning of aggregation protocols P3 and P4. The performance is reported in terms of MCC as functions of the number of training epochs. \modPB{Note that the MCC for training are reported using undersampling so that class imbalance ratio is 2 and that explains the difference in MCC between training vs. test sets. On training sets, with little class imbalance (as it is corrected), the method reports an almost perfect MCC, close to 1 while test set gives the true performance generalized to the much imbalanced data that we necessarily encounter in test conditions. The fact that the MCC can be as high as 0.4 or more indicates a good generalization on the test set despite this imbalance. The values of the TPR and FPR in Table~\ref{tab:confusion_compare} supports also this conclusion.} Also interestingly, it shows that aggregation protocol P3 (Combined-$T_sZ_g$) \emph{learns} faster but \emph{generalizes} less; hence it \emph{overfits} data as compared to aggregation protocol P4 (Stacked-$T_sZ_g$). 
This thus confirms that protocol P4 (Stacked-$T_sZ_g$) performs better in heatwave prediction. 
Note finally that the benefits are more pronounced for the rarest ($1.25\%$) class of extreme events. 
\\
\indent Finally, and importantly, Fig.~\ref{fig:compare} suggests that the transfer learning procedure leads to comparable or better performance, compared to without transfer learning, and that such improved performance is obtained within a single epoch of training, as opposed to the 5 to 10 epochs needed to achieve convergence in performance without transfer learning. 
Transfer learning thus leads to better performance obtained at a significantly decreased computational cost.\\

\begin{table*}[t]
    \centering
    \begin{tabular}{c||c|c|c||c|c|c||c|c|c}
    Class imbalance & \multicolumn{3}{c||}{1/level} & \multicolumn{3}{c||}{10} & \multicolumn{3}{c}{4}\\
    \hline
    Undersampling rate & 1 & 1 & 1 & 1/2 & 1/4 & 1/8 & 1/5 & 1/10 & 1/20\\
    \hline
    Levels & $5\%$ & $2.5\%$ & $1.25\%$ & $5\%$ & $2.5\%$ & $1.25\%$ & $5\%$ & $2.5\%$ & $1.25\%$\\
    \hline
    No transfer learning & & & & & & & & \\
    \hline
    Average MCC & $0.36$ & $0.28$ & $0.23$ & $0.39$ & $\bf 0.39$ & $\bf 0.31$ & $\bf 0.44$ & $0.38$ & $0.10$ \\
    Median MCC & $0.36$ & $0.28$ & $0.23$ & $0.40$ & $0.39$ & $0.33$ & $0.43$ & $0.38$ & $0.04$ \\ 
    Std MCC & $0.03$ & $0.07$ & $0.08$ & $0.04$ & $0.04$ & $0.08$ & $0.04$ & $0.03$ & $0.12$ \\
    \hline
    Transfer learning & & & & & & & & \\
    \hline
    Average MCC & $-$ & $0.27$ & $0.18$ & $-$ & $0.33$ & $0.25$ & $-$ & $0.38$ & $0.31$ \\
    Median MCC & $-$ & $0.25$ & $0.20$ & $-$ & $0.33$ & $0.25$ & $-$ & $0.37$ & $0.32$ \\ 
    Std MCC & $-$ & $0.06$ & $0.06$ & $-$ & $0.06$ & $0.08$ & $-$ & $0.04$ & $0.07$ \\
    \end{tabular}
    
    \begin{tabular}{c||c|c|c||c|c|c}
    Class imbalance & \multicolumn{3}{c||}{2} & \multicolumn{3}{c}{1} \\
    \hline
    Undersampling rate & 1/10 & 1/20 & 1/40 & 1/19 & 1/39 & 1/79 \\
    \hline
    Levels & $5\%$ & $2.5\%$ & $1.25\%$ & $5\%$ & $2.5\%$ & $1.25\%$ \\
    \hline
    No transfer learning & & & & & \\
    \hline
    Average MCC & $\bf 0.44$ & $0.37$ & $0.29$ & $0.41$ & $0.20$ & $0.09$ \\
    Median MCC & $0.44$ & $0.37$ & $0.29$ & $0.42$ & $0.24$ & $0.08$ \\ 
    Std MCC & $0.03$ & $0.03$ & $0.06$ & $0.03$ & $0.10$ & $0.06$ \\
    \hline
    Transfer learning & & & & & \\
    \hline
    Average MCC & $-$ & $\bf 0.40$ & $\bf 0.35$ & $-$ & $0.38$ & $0.34$ \\
    Median MCC & $-$ & $0.40$ & $0.36$ & $-$ & $0.39$ & $0.34$ \\ 
    Std MCC & $-$ & $0.04$ & $0.05$ & $-$ & $0.03$ & $0.04$ \\
    \end{tabular}
    \caption{ \label{tab:undersampling} 
    {\bf Compared performance for heatwave occurrence prediction with different undersampling rates.} Performance is reported in terms of average, median and std MCC, obtained from $40$ independent learning, for aggregation protocol P4 (Stacked-$T_sZ_g$) for the three levels of anomalies, without (top) and with (bottom) transfer learning, for no undersampling (left) to high undersampling (right). This table shows that the undersampling strategy of the large non-extreme event class during the training phase is effective for an imbalance ratio of the order of two.}
\end{table*}

 \begin{figure*}
     \centering
     \includegraphics[width=\linewidth]{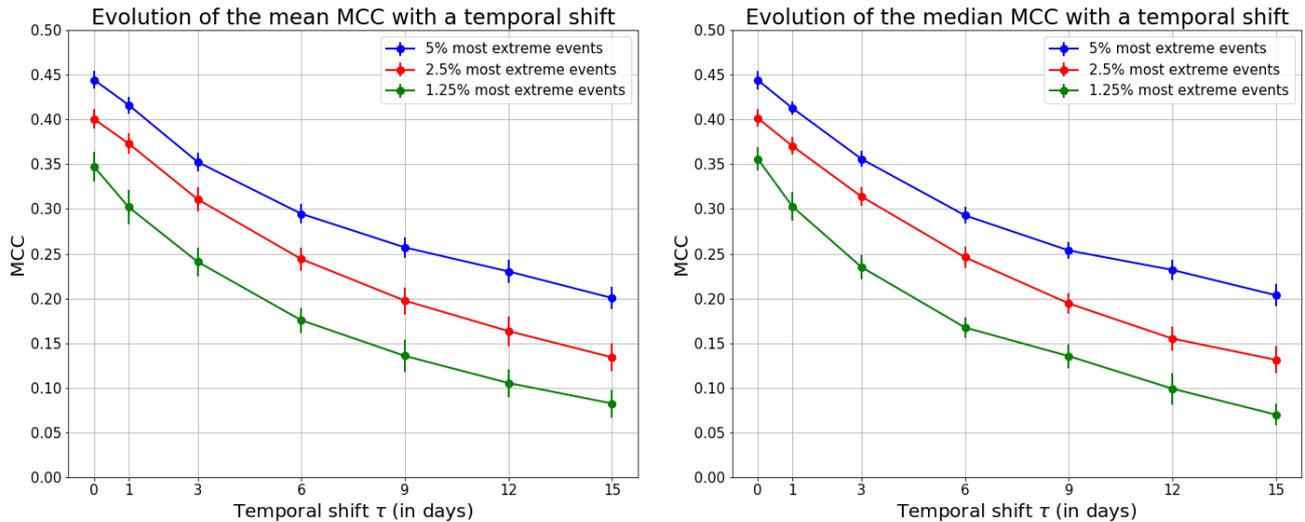}
     \caption{\label{fig:compare_vs_tau} {\bf Heatwave occurrence prediction performance in MCC as function of the number of days in advance $\tau$.} Mean MCC, with standard deviation,   (left plot) and median MCC, with max absolute deviation,  (right plot), obtained as averages across the $40$ independent learning, for the three levels of extreme events. These plots clearly show that achieved MCC are significantly above $0$ indicating the ability of the proposed Deep Learning procedure to predict heatwave extreme events $\tau$ days in advances from the sole observations of the surface temperature and geopotential height fields at a single date.}
 \end{figure*}

 \begin{figure*}
     \centering
     \includegraphics[width=\linewidth]{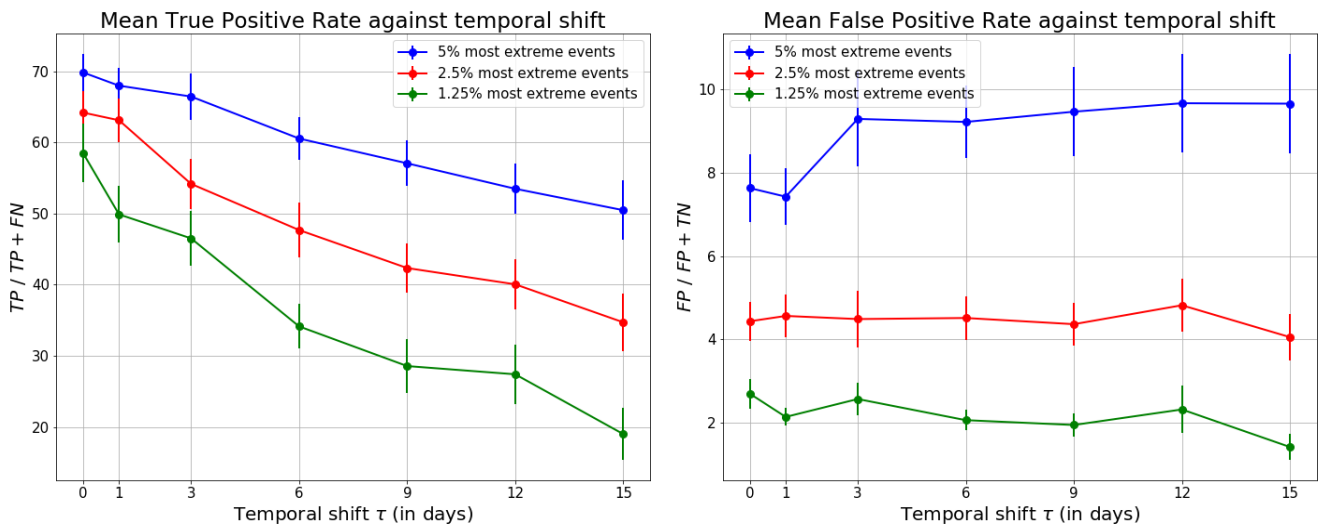}
     \caption{\label{fig:compare_TP_FP} 
     {\bf Compared performance for heatwave occurrence prediction as a function of $\tau$.} Compared rates (in percentages) of True positive (left plot) and False positive (right plot) predictions, for each heatwave levels ($5\%, 2.5\%$ and $1.25\%)$. Average across $40$ independent learning as described in Figure \ref{fig:compare}. Percentages of True Positives (resp. False Positives) are quantified with respect to the sizes of the positive (resp. negative) class.}
 \end{figure*}

\noindent {\bf Undersampling rate.} In general, supervised learning (and a fortiori deep learning) for forecasting of extreme events faces potentially severe class imbalance.
As described in Sec.~\ref{sec:dl}, \modPB{it has been chosen to address this issue during training by undersampling the large class of non-extreme events in the training set.}
Table~\ref{tab:undersampling} compares achieved performance in terms of average MCC for different imbalance ratios between the non-extreme and extreme class size, \modPB{varying from $1$ (undersampling so that we have equal class size), to $2$, $4$ and $10$, and performance obtained without undersampling (so that we have 19 non-extreme events for 1 heatwaves at level of 5\%)}.
Table~\ref{tab:undersampling} shows that the undersampling strategy of the large non-extreme event class during the training phase is effective as soon as it brings the training class-size imbalance to a ratio of $1$ or $2$, while performance degrades with larger ratios, $4$ and $10$, or no undersampling -- hence large class-imbalance.
This is particularly clear with transfer learning. 
To achieve optimal prediction performance, it is thus not mandatory that classes have exactly the same size in the training set, but it is critical that class imbalance ratio remains limited of a few units. \\
\indent For the sake of completeness, let us mention that the discussions related to data aggregation protocols and transfer learning, were presented with a class imbalance ratio of $2$ corresponding to sampling rate of $1/10$, $1/20$ and $1/40$ for the three levels of extreme events. 
Equivalent conclusions were drawn from analyzing results obtained with a class-size ratio of $1$ corresponding to sampling rate of $1/19$, $1/39$ and $1/79$.

\subsection{Forecasting performance}
\label{sec:resultsb}

\noindent {\bf Heatwave prediction scheme.} The methodological analyses reported above in Table~\ref{tab:scores} already yield the first key result of the present article: 
the occurrence of heatwaves can be predicted with some success for the three levels of extreme events within the next $D$ days from present time $t$, from data observed across space at the sole time $t$, and this for the four aggregation protocols. 

Further, these analyses show the better performance of the aggregation protocol P4 (Stacked-$T_sZ_g$), with transfer learning and non-extreme event class undersampling rates of $1/10$, $1/20$ and $1/40$ for the three levels of heatwaves, as optimal. 
These findings permit to conduct now a systematic analysis of prediction performance as functions of $\tau$, the number of days in advance heatwaves have to be predicted: that is, surface temperature and geopotential height fields at date $t-\tau$ are used to predict a heatwave occurring any time between dates $t$ and $t+D$. 

The training procedure is repeated $40$ times from scratch with independent yearly-based train/test data split, as for $\tau =0$ and as described in Section~\ref{sec:dl}.\\

\noindent {\bf Heatwave prediction performance.} 
Fig.~\ref{fig:compare_vs_tau} reports, for the three level of extreme events, the means, $\pm $ standard deviations (left), and medians $\pm$ max absolute deviations (right), obtained as averages across the $40$ independent learning, as functions of the prediction delay $\tau$. 

These plots demonstrate that the achieved MCC is significantly positive for the three levels of extreme events and for $0 \leq \tau \leq 15$ days. 
They also show that the decrease in MCC as function of $\tau$ is slow, decreasing from the range $0.35$ to $0.45$ at $\tau = 0 $ to the still significant range of $0.10$ to $0.20$ at $\tau = 15 $, thus that occurrence prediction as early as $15$ days in advance is achieved.


To complement these performance analyses, Fig. \ref{fig:compare_TP_FP} reports the evolution, with respect to the prediction scale $\tau$, of the TPR and FPR (i.e., respectively  $TP/(TP+FN)$ and $FP/(FP+TN)$, in percentage). 
It shows that the decrease in performance quantified by the decrease of MCC in Fig~\ref{fig:compare} stems from a decrease in the TPR:
While the prediction horizon $\tau$ increases, fewer heatwaves are detected. 
The False Positive Rate remains constant (and hence so does the number of FP) to less than $10\%$ of the total number of negative samples: 
While the prediction horizon $\tau$ increases, the detection of negative events remains "as easy".


Altogether, achieved performance yield the following conclusions, consisting per se of relevant findings for climatologists and for ML practioners tackling this challenging task: \\
\indent i) The surface temperature and geopotential height spatial fields in North hemisphere at a single observation time contain sufficient spatial structures and information to predict the occurrence of heatwaves over European territory of the size of France, up to 15 days ahead of the beginning of a long-lasting heat wave. \\
\indent ii) Beyond independent spatial-dynamics, the cross-spatial dynamics of these two fields contain relevant structures permitting to enhance significantly prediction performance. 
The fact that the aggregation protocol P4 (Stacked-$T_sZ_g$) outperforms other field combination strategies clearly indicates that such cross-dynamics must be processed jointly from the finest available physical scales.\\
\indent iii) CNN-layer based deep-learning architectures are able to extract relevant (cross-)spatial dynamics of climate data. 
Convolutional filter sizes were varied and results were reported here only for the best prediction performance, corresponding to filter size ranging from $9 \times 9$ to $12 \times 12$. 
In physical units, this corresponds to filters exploring jointly frequency bands of width $\Delta f \sim 10^{-3}\text{km}^{-1}$ and thus (cross-)spatial dynamics within territories of size roughly corresponding to $1000 \times 1000$ km$^2$. 
Incidentally, this turns out to correspond to the size of a typical spatial correlation length, the order of magnitude of the size of cyclonic and anticyclonic anomalies, of the order of the Rossby deformation radius \cite{vallis2017atmospheric}.\\
\indent iv) Predicting heatwaves at $\tau=0$ is already an impressive outcome since it corresponds to predicting the occurrence of an extreme event, within the next $D=14$ days from the observation of a extremely limited amount of climate data potentially available for prediction (2 spatial fields only at a single observation time). 
The ability of the proposed scheme to predict heatwaves as early as $15$ days in advance is even more impressive.
Indeed, typical correlation times in climate time series are documented to be of 3 to 5 days \cite{vallis2017atmospheric}. 
Predicting the occurrence of heatwaves $3$ to $5$ times ahead of that correlation time suggests that the proposed forecasting scheme has extracted relevant fine (cross-)spatial structures from data, a remarkable outcome.

\section{Discussion and perspectives}
\label{sec:conclusion}

The present work has illustrated and quantified the ability of deep learning approaches to predict the forthcoming occurrence of long-lasting heatwaves, from 1000-year of a climate model output. 

One key result is that significant prediction performance can be achieved from the analysis of the (cross-) spatial dynamics of only two fields, the surface temperature and 500 hPa geopotential height, observed at a single time. The forecast gives significant results for time ahead which are much larger that the field correlation time scales.

These successes are grounded: \\
i) on the ability to use a large size training database, consisting here of $1000$ years of simulated climate data, as well as the use of surface temperature and geopotential height, chosen a priori as relevant information to heatwave dynamics from the existing scientific literature;\\
ii) on the use of CNN-layer based neural network architectures;\\
iii) on combining CNN with in-depths analyses of issues such as data aggregation, non-extreme event large class undersampling to address class-size imbalance intrinsically associated with extreme-event predictions, transfer learning and nested extreme event structure to achieve relevant prediction of the most extreme events, using learning performed from less extreme events.
The study and assessment of these three practical procedures can be seen as methodological contributions valid in generic settings and in other applications facing extreme event predictions and imbalanced class sizes. 

At the application level, the claim is not that deep learning approaches should replace physics-driven models in climate predictions. 
Rather, the present work can be read as a proof-of-concept result for the use of learning procedure and, here, a specific deep learning architecture, in climate extreme event predictions. 
At this stage, it mostly provides climatologist with a \emph{black-box} tool that performs heatwave occurrence predictions with satisfactory performance, and at very low computational costs in time and computer resources and using very limited sets of observations.
Achieving the same task with the traditional physics-model based approach requires solving a set of dynamical partial differential equations in climate simulator engines, involving significant computational resources and observed data for initialisation. 

From the point of view of atmosphere dynamics, the key result of this paper is that our machine learning approach has significant forecast skills for long-lasting heat waves up to 15 days ahead of the beginning of the event. This predictability range is long, compared to what might have been expected, for the dynamics of midlatitude atmosphere. A very interesting perspective, that goes beyond the scope of this work, would be to identify the dynamical mechanisms at the core of this potential predictability. \FB{For this aim, it would be useful to try to incorporate physics knowledge and interpretability of the trained neural network to contribute to the understanding of the physical mechanisms at work in heatwave dynamics.}

\FB{While the present is the first to use deep neural network to forecast long-lasting extreme heat waves, an analogous approach has been used recently \cite{pedram} to forecast intra-day heat waves. Because the phenomenologies of long lasting and intra-day heat wave are very different, it does not make much sense to compare directly the predictive skills of the two approaches. However it would interesting to consider in the future if CapsNet used in \cite{pedram} might improve our approach, and if the transfer learning and class imbalance used in the present study might improve the prediction of intra-day heatwaves in \cite{pedram}.}

This work will be continued along several lines.
On the application side, the extent to which aggregating other spatial fields available (e.g., using the geopotential height for several values of atmospheric pressure) on the same day would improve prediction performance will be investigated.
Also, it will be studied how combining observations made across several times, thus aggregating temporal and spatial dynamics in deep learning architectures, can be done to improve heatwave occurrence forecasting performance. Further, the prediction of shorter-duration heatwaves, or of heatwaves occurring on different areas will be analyzed. 



At the methodological level, it would be natural to try to relate prediction performance to architecture complexity, which could be quantified using the Vapnik-Chervonenkis Dimension tool, as recently suggested and explored in \cite{baum1989size,friedland2017capacity,liotet2020deep}. 

We now come back to the three key issues in the study of rare climate extreme events: lack of historical data, difficulty to sample extremely rare events with models, and the assessment of model biases for extreme in climate models.  

\PA{Regarding} the lack of historical data, \PA{an} interesting perspective would be to connect \PA{the proposed} machine learning approach for extreme long-lasting heat waves to observation or reanalysis data. We first note that the predictive value of \PA{the proposed} approach drops very fast if we use much less than 1 000 years of data for training (not shown). This means that a deep neural network, with the same predictive capability as ours, most probably cannot be trained using only 70 years of the available reanalysis data. This statement seems obvious if we deal with unprecedented events, never observed in the dataset. The use of observation or reanalysis datasets would anyway be very interesting, but it should necessarily be coupled in an indirect way with other datasets produced by climate model or weather forecast systems, for instance through transfer learning. This is a very interesting perspective for future works. 

\FB{Regarding the difficulty of sampling exceptionally rare extreme events, for instance unprecedented extreme heat waves, we have recently developed rare event simulation techniques that are able to multiply by several orders of magnitude the number of observed heat waves with PLASIM model \cite{ragone2019} and with CESM (the NCAR model used for CMIP experiments) \cite{ragone2021}. We are currently working on coupling these rare event simulations with the machine learning forecast developed in this paper. The point is to improve both rare event simulations using machine learning forecast, and machine learning forecast using the unprecedented heat wave statistics obtained with rare event simulations. This is a interesting perspective to propose solutions to the key fundamental issue that is the lack of data in the science of climate extremes.} 

\section{Additional Requirements}

\section*{Conflict of Interest Statement}

The authors declare that the research was conducted in the absence of any commercial or financial relationships that could be construed as a potential conflict of interest.

\section*{Author Contributions}

Patrice Abry, Pierre Borgnat and Freddy Bouchet have devised the scientific project. Francesco Ragone has produced the Plasim model data. Valerian Jacques-Dumas has implemented and run the machine learning studies and prepared the figures. All the authors have equally participated to the analysis, thinking, and evolution of the scientific project, and have equally contributed to the writing of the paper.

\section*{Funding}

This work was upported by the  ACADEMICS Grant of IDEXLYON, Univ. Lyon, PIA ANR-16-IDEX-0005. This work was support by the ANR grant SAMPRACE (F. Bouchet), project ANR-20-CE01-0008-01. The computation of this work were partially performed on the PSMN platform and the CBP center of ENS de Lyon. This work was granted access to the HPC resources of CINES under the DARI allocations A0050110575,  A0070110575 and A0090110575 made by GENCI.

\section*{Acknowledgments}
We thank Bastien Cozian, Dario Lucente, George Miloshevich for useful comments on this work. 


\section*{Data Availability Statement}
The data analyzed in this study was model outputs from the PlaSim model, for an existing $1000$-year simulation, of a climate which is characteristics of the $90$'s. This dataset might be shared upon request to Freddy Bouchet. 

\nocite{*}
\bibliographystyle{frontiersinHLTH&FPHY} 
\bibliography{Article}

\end{document}